%% file: root.tex
\documentclass[letterpaper, 10 pt, conference]{ieeeconf}  

\IEEEoverridecommandlockouts                              

\usepackage{subfiles}
\usepackage{amsmath}
\usepackage{amssymb}
\usepackage{mathtools}
\usepackage{graphicx}
\usepackage{float}
\usepackage{subcaption}
\usepackage{epstopdf}
\usepackage[tableposition=top]{caption}
\usepackage{array}
\usepackage{url}
\usepackage[ruled,lined]{algorithm2e}
\usepackage{setspace}
\usepackage{algpseudocode}
\usepackage{diagbox}
\usepackage{multirow}
\usepackage[table]{xcolor}
\usepackage[framemethod=TikZ]{mdframed}
\usepackage[procnames]{listings}
\usepackage[hidelinks]{hyperref}

\mdfdefinestyle{frameStyle}{%
    roundcorner=5pt,
    backgroundcolor=white}

\newtheorem{definition}{Definition}

\algtext*{EndIf}
\algtext*{EndFor}
\algtext*{EndWhile}
\algtext*{EndFunction}

\renewcommand{\vec}[1]{\mathbf{#1}}

\title{\LARGE \bf
Personalised Robot Behaviour Modelling for Robot-Assisted Therapy\\in the Context of Autism Spectrum Disorder
}

\author{Michał Stolarz$^{\dagger\mathsection}$, Alex Mitrevski$^{\dagger\mathsection}$, Mohammad Wasil$^{\dagger}$, and Paul G. Pl{\"o}ger$^{\dagger}$
\thanks{$^{\dagger}$The authors are with the Autonomous Systems Group, Department of Computer Science, Hochschule Bonn-Rhein-Sieg, Sankt Augustin, Germany,
        {\tt\scriptsize michal.stolarz@smail.inf.h-brs.de, <aleksandar.mitrevski,mohammad.wasil,paul.ploeger>@h-brs.de}} %
\thanks{$^{\mathsection}$Corresponding author} %
}

\begin{document}

\maketitle
\thispagestyle{empty}
\pagestyle{empty}


\begin{abstract}
In robot-assisted therapy for individuals with Autism Spectrum Disorder, the workload of therapists during a therapeutic session is increased if they have to control the robot manually.
To allow therapists to focus on the interaction with the person instead, the robot should be more autonomous, namely it should be able to interpret the person's state and continuously adapt its actions according to their behaviour.
In this paper, we develop a personalised robot behaviour model that can be used in the robot decision-making process during an activity; this behaviour model is trained with the help of a user model that has been learned from real interaction data.
We use Q-learning for this task, such that the results demonstrate that the policy requires about 10,000 iterations to converge.
We thus investigate policy transfer for improving the convergence speed; we show that this is a feasible solution, but an inappropriate initial policy can lead to a suboptimal final return.
\end{abstract}


\subfile{subfiles/intro}

\subfile{subfiles/related_work}

\subfile{subfiles/imitation_game}

\subfile{subfiles/behaviour_personalisation}

\subfile{subfiles/evaluation}

\subfile{subfiles/discussion}

\addtolength{\textheight}{-12cm}   


\section*{ACKNOWLEDGMENT}
This work is conducted in the context of the MigrAVE project, which is funded by the German Ministry of Education and Research (BMBF).
We hereby thank our partners, M{\"u}nster University of Applied Sciences (FHM) and the RFH - University of Applied Sciences, Cologne.
The research in the MigrAVE project has received an ethics approval by the German Psychological Society (DGPs).

\bibliographystyle{IEEEtran}
\bibliography{references}

\end{document}

%% file: subfiles/intro.tex
    \section{INTRODUCTION}
    \label{sec:introduction}

    One of the objectives of robot-assisted therapy (RAT)~\cite{esteban2017build} is increasing the autonomy of the robot that is used during therapy sessions; this has the purpose of reducing the necessary therapist interactions with the robot \cite{david2018developing,robins2017developing,rudovic2017measuring,marinoiu20183d}, while still keeping the therapist in control of the sessions at all times.
    Particularly in the treatment of children with Autism Spectrum Disorder (ASD), RAT focuses on using a robot to facilitate and guide the learning of concepts that affected individuals require in their everyday lives, such as repeating everyday motions or recognising emotions.
    Robots are attractive for this problem because children have been shown to find it more comfortable to interact with a robot than with another person \cite{robins2006}.
    In the context of RAT for children with ASD, robot programs are usually developed in such a way that they can be used generically for different children; however, children with ASD may have different reactions to specific stimuli and, depending on their development, may also benefit from therapy sessions focusing on specific aspects.
    This means that a generic RAT approach may not be optimal for effective treatment of individuals; instead, the robot should be able to adapt its behaviour to the needs of each individual and therapy session \cite{esteban2017build,scassellati2018improving,rudovic2018personalized}.

    This type of adaptation, also referred to as \emph{personalisation}, requires a robot to modify its behaviour to each individual user or to groups of similar users.
    A personalised behaviour model can be learned by involving a user in the learning loop, which is referred to as interactive machine learning (IML) \cite{senft2019teaching}.
    There are two primary types of IML in the context of personalisation, namely \emph{learning from user feedback} and \emph{guidance-based learning}, where the former relies on direct or indirect user feedback, while the latter incorporates feedback from an external observer, for instance a therapist.
    Learning from user feedback can be difficult to perform efficiently because the robot needs to perform exploration to find an appropriate behaviour policy, while guidance-based learning avoids incorrect actions being performed by the robot during the learning process, but at the cost of requiring a supervisor to be involved throughout.
    One way in which the amount of involvement of a user or a supervisor can be reduced is by incorporating a user model \cite{rossi2017user} in the policy learning process, based on which users are represented by particular parameters, such as their engagement.

    In this work, we present a personalised behaviour model that (i) is based on the concept of learning from feedback and (ii) incorporates learned user models that estimate a user's engagement and expected performance in an activity.
    We represent both engagement and performance by Gaussian processes.
    These are learned from data collected from multiple users, such that users are split into clusters and a dedicated user model is learned for each cluster.
    Based on these models, we learn a policy that a robot uses for selecting the difficulty level of an activity and the type of provided feedback to the user.
    We compare different rewards for the policy learning algorithm and investigate a policy pretraining method for accelerating the policy convergence speed.
    In this paper, we evaluate the feasibility of the proposed method through an experiment in which adults without ASD\footnote{Before using the proposed method with children with ASD, we were interested in evaluating whether the method works in principle; this is the main purpose of the experiments in this paper.} were playing an emotion sequence memorisation game with QTrobot \cite{luxai2017qtrobot}.
    The results of the experiments demonstrate that the models can be useful for personalising the robot's behaviour to different users, but dedicated experiments with children with ASD are needed to evaluate the usefulness of the method in that context as well.

%% file: subfiles/related_work.tex
    \section{RELATED WORK}
    \label{sec:related_work}

    Personalised behaviour models can be categorised into various personalisation dimensions \cite{tsiakas2018taxonomy}, but we find two particular dimensions that are useful for RAT for children with ASD \cite{stolarz2022personalized}, namely social behaviour and activity difficulty personalisation.
    This means that the robot should be able to (i) react to a user's disengagement during an activity and (ii) adjust the activity difficulty to each user.
    In this section, we briefly discuss work dealing with these two types of personalisation; some of this discussion is based on \cite{stolarz2022personalized}.

    Social behaviour personalisation is a technique used for maintaining a user's interest in the interaction by making the robot provide appropriate verbal and nonverbal expressions to the user, such as gestures.
    One technique in this context is the supervised autonomy system in \cite{esteban2017build,cao2019robot}, where the robot adapts its behaviour to the behaviour of children with ASD.
    The learning mechanism is similar to learning from guidance, as the robot asks a therapist for feedback before executing any actions; however, learning is performed on data from all participants and is thus not personalised.
    For behaviour personalisation, a feed-forward network is used in \cite{senft2015sparc}, but this can be inappropriate for online real-time interaction in long-time scenarios, as the learning time increases with the amount of collected data.
    Another approach is based on the Q-learning algorithm \cite{watkins1992q,senft2017supervised}, which, on its own, requires a significant number of interactions with the user before convergence; to accelerate the convergence, the problem has to be decomposed so that the Q-value table stays relatively small \cite{hemminahaus2017towards}.
    Most of these approaches are based on learning from guidance \cite{esteban2017build,senft2019teaching,senft2015sparc,senft2017supervised}, which avoids the risk that the robot performs inappropriate actions during learning; on the other hand, when applying learning from feedback \cite{hemminahaus2017towards,chan2012social}, the robot has to explore actions on its own, which may be undesired in ASD therapy.

    To maintain a user's interest in the interaction, it is also important to personalise the difficulty of an \emph{activity}\footnote{Here, the term activity refers to a task that is to be performed by the user, such as memorising a sequence or a therapeutic exercise.} so that it matches each child's skill level during RAT.
    Most existing techniques that deal with this type of personalisation are based on learning from feedback.
    In \cite{jain2020modeling,clabaugh2019long}, reinforcement learning (RL) is used for personalising the feedback and instruction difficulty levels during math games; however, the behaviour models used here do not include reactions that directly control for the engagement of a user during an interaction.
    An approach for both social behaviour personalisation and game difficulty personalisation is presented in \cite{tsiakas2018task}.
    Here, a policy is learned using Q-learning and various methods of updating the Q-table are explored to influence the policy convergence speed; an approach for modelling human users is introduced as well, which has the purpose of making the training of the framework more practically feasible.
    A technique for increasing the learning speed is also proposed in \cite{tsiakas2016adaptive}, where the idea is to initialise the policy to be learned with a policy that has been trained with a certain user model, but this only personalises the game difficulty.

    Our work is primarily based on \cite{jain2020modeling,tsiakas2018task,tsiakas2016adaptive}, but, to address some of the aforementioned deficits, we (i) reduce the size of the state space to simplify the behaviour model learning problem, (ii) use a Gaussian process for user modelling, and (iii) estimate engagement using facial features.\footnote{Compared to \cite{tsiakas2018task}, where a headset is used for engagement estimation, and in line with \cite{jain2020modeling}, we believe that an external engagement observer is more convenient for children with ASD, as they may be overwhelmed by the sensory stimuli if they need to wear an additional device \cite{javed2019robotic}.}

%% file: subfiles/imitation_game.tex
    \section{ROBOT-ASSISTED GAME}
    \label{sec:robot_assisted_game}

    To ground the personalised behaviour model that is presented in this paper to a concrete task, we use a game whose objective is to evaluate the ability of users to memorise and repeat sequences of spoken emotions; a game of this type can target the social skills of children with ASD and has also been used in \cite{scassellati2018improving}.
    Our game is designed based on \cite{tsiakas2018task}, such that each user session consists of $\omega = 10$ sequences to memorise.
    Each sequence $\Omega_j, 1 \leq j \leq \omega$ consists of words that are randomly sampled from a pool of four emotions, namely $\{$happy, disgusted, sad, angry$\}$; a sequence can have a length $|\Omega_j|$ of 3, 5 or 7 emotions, with respective difficulty levels $L_j \in \{1, 2, 3\}$.
    During the game, the robot says each $\Omega_j$ out loud and the user has to reproduce the sequence by selecting images corresponding to the emotions on a tablet.
    To reproduce a sequence correctly, the user has to choose the correct image for every emotion in $\Omega_j$ in the right order.
    During the game, the robot should choose sequence lengths $|\Omega_j|$ that are appropriate for the user and provide feedback $F_j$ so that they remain engaged in the interaction; thus, we want the selection of robot actions to be based on the user's game performance and engagement level.

    We use QTrobot \cite{luxai2017qtrobot} as a robotic assistant in this work, which is a robot developed for tablet-based therapeutic games.
    The robot has an Intel RealSense D435 depth camera, a Raspberry Pi, and also includes an Intel NUC PC for more demanding computations.
    QTrobot is integrated with two tablets --- one for the therapist and one for the child in therapy; this allows therapists to control the robot or start appropriate games during a session, while the child tablet is only supposed to execute the games chosen by the therapist.

%% file: subfiles/behaviour_personalisation.tex
    \section{BEHAVIOUR PERSONALISATION}
    \label{sec:behaviour_personalisation}

    The objective of this work is to develop a personalisation strategy for RAT, with a particular focus on adapting the behaviour of a robot in terms of controlling the activity difficulty and providing appropriate user feedback.
    For this purpose, we present a method for learning a robot behaviour policy based on user feedback that incorporates a learned user model in the policy learning loop.
    We particularly discuss the user modelling and the policy learning process, and also investigate policy pretraining for speeding up the policy convergence.

    As discussed in section \ref{sec:introduction}, the purpose of using a user model is to reduce the amount of user or therapist interactions that are needed for learning a behaviour policy.
    We utilise a user model that estimates the engagement and expected performance of a group of similar users in a given activity, such that we assume that both performance and engagement are represented by numerical values.
    \begin{mdframed}[style=frameStyle]
        \begin{definition}
            A user model $\mathcal{M}$ is a tuple $\mathcal{M} = (F^p, F^e)$, where $F^p$ is a performance prediction component and $F^e$ is an engagement estimation component.
        \end{definition}
    \end{mdframed}
    A model $\mathcal{M}$ is learned from user data collected during real interactions, where the estimated engagement and activity performance are recorded.
    These data are then clustered in order to identify groups $C_k, 1 \leq k \leq c$ of similar users, namely users that have similar performance and engagement during an activity, where $c$ is the number of user groups.
    For each $C_k$, both user model components are learned from the data as Gaussian processes (GPs) \cite{gp_book}, which have the desirable property of encoding prediction uncertainty.
    The learned user model is then incorporated into a policy learning loop, such that the policy $\pi$ is learned using Q-learning on discrete state and action spaces.
    In the rest of this section, we describe the engagement estimation, the model $\mathcal{M}$, and the policy learning in more detail.

    \subsection{Engagement Estimation}
    \label{subsec:engagement_estimation}

    To estimate the engagement of a user during an activity, we use a binary classifier $\mathcal{E}: \mathbb{R}^{32} \rightarrow \{-1, 1\}$ based on~\cite{jain2020modeling}; here, $1$ denotes engagement, namely that the participant is actively involved in the interaction and pays attention to the robot, while $-1$ denotes disengagement, namely that the participant is not focused on the robot.
    For the classifier, we use 32 features --- head pose (6 features), facial action units (18 features), and gaze position and angle (8 features); these are extracted using the OpenFace library \cite{baltrusaitis2018openface}.

    To collect training data for the classifier, we asked participants to act out the aforementioned engagement and disengagement criteria; this has the potential disadvantage that the participants' behaviour may not be completely natural during the interaction, but simplifies the data labelling effort since the interactions are appropriately segmented during data collection.
    Using the training data, we performed an evaluation procedure similar to \cite{jain2020modeling} to select a suitable classifier.
    We particularly performed leave-one-out cross validation and compared multiple classifier types.
    Based on this evaluation, we use an XGBoost\footnote{\url{https://xgboost.readthedocs.io}} classifier in this study, which has a validation accuracy of about $85\%$.\footnote{The training implementation can be found at \url{https://github.com/migrave/migrave_models}}

    \subsection{User Model}
    \label{subsec:user_model}

    Given a dataset $X$ for $n$ users, where the activity scores, solved difficulty levels, estimated engagement values, and robot feedback types were recorded throughout the activity, we (i) represent each user as a vector, (ii) perform dimensionality reduction to 2D space, (iii) cluster the users into $c$ groups, and (iv) train $c$ models $\mathcal{M}_k$ (one for each user cluster) to predict activity performance and engagement scores for unseen activity states.
    We represent each user $U_j, 1 \leq j \leq n$ by a vector $\vec{u}_j$
    \begin{equation}
        \vec{u}_j = \left(p_{j,1:l}, e_{j,1:l}\right)
        \label{eq:user_vector}
    \end{equation}
    where $p_{j,1:l}$ are the success probabilities of solving each sequence difficulty level $L_i, 1 \leq i \leq l$ and $e_{j, 1:l}$ is the user's mean engagement score for the respective difficulty levels.\footnote{This mean is calculated only for periods when the user is supposed to be focused on the robot, for example when the robot is talking to the user.}
    We then project $\vec{u}_j$ for all $n$ users onto a 2D space using PCA and apply K-means clustering to group the projected vectors into $c$ clusters.
    Given the assignment of users $U_j$ to clusters $C_k$, we create a performance model $F^p_k$ and engagement model $F^e_k$; these comprise the user model $\mathcal{M}_k$ for cluster $C_k$.
    Both $F^p_k$ and $F^e_k$ are GPs that can be used for regression to unobserved states
    \begin{align}
        F^p(\vec{s}^p) &= GP\left(\mu(\vec{s}^p), k(\vec{s}^p, \vec{s}^{p}{'})\right) \\
        F^e(\vec{s}^e) &= GP\left(\mu(\vec{s}^e), k(\vec{s}^e, \vec{s}^{e}{'})\right)
    \end{align}
    where $\mu$ is the mean and $k$ the covariance of the inputs.

    $F^p_k$ predicts how likely the users in $C_k$ are to succeed in a given activity state $\vec{s}^p$, namely $F^p_k(\vec{s}^p) \mapsto [0, 1]$, where each $\vec{s}^p = (L, F, PS)$.
    Here, $L \in \{1,...,l\}$ is the current difficulty level, $F \in \{0,1,2\}$ is the given robot feedback (no feedback, encouraging feedback, or challenging feedback, respectively), and $PS \in \{-l, -l+1, ..., 0,..., l-1, l\}$ is the activity score achieved by the user in the last sequence.\footnote{A positive or negative score is given for a correctly or incorrectly solved sequence of a given difficulty level, respectively, and a $0$ score is only used in the initial state, before a sequence length has been selected.}

    $F^e_k$ estimates the expected engagement value for the users in $C_k$, namely $F^e_k(\vec{s}^e) \mapsto [-1, 1]$, where $\vec{s}^e = (L, F, PS, O)$.
    Here, $O \in \{1, -1\}$ stands for the outcome (correct or wrong) of solving the current sequence.

    \subsection{Robot Behaviour Model}
    \label{subsec:robot_behaviour_model}

    Our behaviour model for robot decision making is represented as a discrete Markov Decision Process $\mathcal{B} = (S, A, P_a, R_a, \gamma)$, where $\gamma$ is a discount factor, each state $\vec{s} \in S$ is defined as $\vec{s} = (L, F, PS)$ and the action space $A$ consists of actions $a_i, 1 \leq i \leq l+2$; here, $a_{1:l}$ are used to set a difficulty level $L$ for the next sequence, while $a_{l+1}$ and $a_{l+2}$ say either encouraging ($a_{l+1}$) or challenging feedback ($a_{l+2}$) and repeat the same $L$ for the next sequence.\footnote{It should be noted that actions $a_{l+1}, a_{l+2}$ cannot be performed for the first sequence in a session as there is no context to give feedback.}
    The robot in state $\vec{s}_t$ moves to a state $\vec{s}_{t+1}$ after performing action $a$ with probability $P_a(\vec{s}_{t+1}|\vec{s}_t)$ and receives an immediate reward $R_a(\vec{s}_t)$, which can be based on (i) the activity result $RE \in \{-1, 1, ..., l\}$, where $-1$ is given for a wrong answer and $1,...,l$ for a correctly solved sequence of level $L$, and (ii) the mean engagement score $E \in [-1, 1]$ calculated after the user solves a given sequence.\footnote{$RE$ is assigned $-1$ for an incorrect answer regardless of the difficulty level $L$ in order not to discourage the robot from choosing a high $L$.}

    We perform model learning with the tabular Q-learning RL technique, which is summarised in Alg. \ref{model_learning}. Here, $Q(\vec{s}_t, a_t)$ is the value of a given entry in the Q-value table, $\vec{s}_t$ and $\vec{s}_{t+1}$ are the states before and after the execution of $a_t$, respectively, and $R_{a}(\vec{s}_t)$ is the immediate reward after applying $a_t$ in $\vec{s}_t$.
    $R_{a}(\vec{s}_t)$ is calculated by a function $F_r$, such that we experiment with different functions in the evaluation.
    The action $a_t$ is selected with the softmax exploration strategy, namely there is a unique temperature parameter $T_{\vec{s}_t}$ for each state $\vec{s}_t$, which is decreased according to the number of visits to $\vec{s}_t$.
    Finally, $\alpha$ is a predefined learning rate.

    \begin{algorithm}
        \begin{algorithmic}[1]
            \Function{\texttt{Q-iteration}}{$t, k, l, Q, \gamma, \alpha, \vec{s}_t, T_{\vec{s}_t}, CS_t$}
                \State $a_t \sim \frac{e^{Q(\vec{s}_t, a_t)/ T_{\vec{s}_t}}}{\sum_{i=1}^{l+2}e^{Q(\vec{s}_t, a_i)/ T_{\vec{s}_t}}}$
                \State $L_{t+1}, F_{t+1} \leftarrow a_t(\vec{s}_t)$
                \State $PS_{t+1} \leftarrow CS_t$
                \State $\vec{s}_{t+1} \leftarrow (L_{t+1},F_{t+1},PS_{t+1})$
                \State $p(\text{success}|\vec{s}_{t+1}) \leftarrow F^p_k(\vec{s}^{p}_{t+1})$
                \State $O_{t+1} \leftarrow \begin{cases} 1 & p(\text{success}|\vec{s}_{t+1}) \geq \mathcal{U}(0,1) \\ -1 & \text{otherwise} \end{cases}$
                \State $RE_{t+1} \leftarrow \begin{cases} L_{t+1} & \text{if $O_{t+1}=1$} \\ -1 & \text{if $O_{t+1}=-1$} \end{cases}$
                \State $E_{t+1} \leftarrow F^e_k(\vec{s}^{e}_{t+1})$
                \State $R_a(\vec{s}_t) \leftarrow F_r(RE_{t+1}, E_{t+1})$
                \State $Q(\vec{s}_t, a_t) \leftarrow Q(\vec{s}_t, a_t) + \alpha(R_a(\vec{s}_t) + \gamma\max\limits_a Q(\vec{s}_{t+1}, a) - Q(\vec{s}_t,a_t))$
                \State $CS_{t+1} \leftarrow L_{t+1}\cdot O_{t+1}$
                \State \Return $Q$
            \EndFunction
        \end{algorithmic}
        \caption{One loop iteration of the learning procedure of $\mathcal{B}$. Here, $t$ is the current time step, $CS_t$ the current user score, and $\mathcal{U}$ a uniform distribution.}
        \label{model_learning}
    \end{algorithm}

%% file: subfiles/evaluation.tex
    \section{EVALUATION}
    \label{sec:experiments}

    \subsection{Experimental Setup}
    \label{sec:experiments:setup}

\begin{figure}
        \centering
        \includegraphics[width=0.65\linewidth]{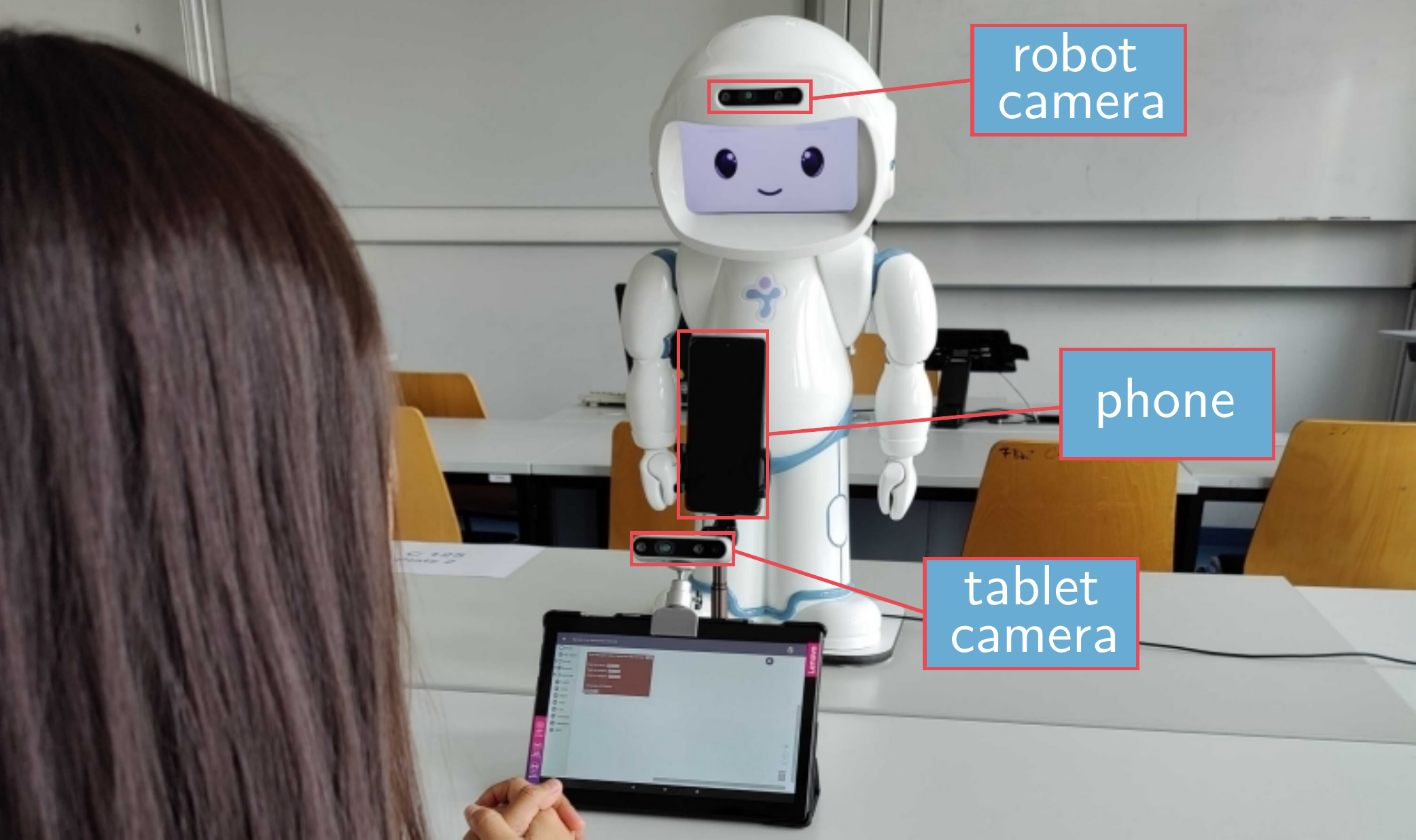}
        \caption{Setup for the sequence learning game}
        \label{fig:experiment_setup}
    \end{figure}

    To evaluate the feasibility of the proposed user models and robot behaviour model, we collected data from 20 adult participants who played the game described in section \ref{sec:robot_assisted_game}.
    The setup is depicted in Fig. \ref{fig:experiment_setup}.
    The participants were master's students and research staff members; all of them had prior experience with robots, but some of them had never interacted with QTrobot.
    During the experiment, data were recorded from the robot's sensors, a camera placed on top of the user tablet, and a smartphone placed between QTrobot and the tablet; for this work, we only used data recorded from the robot's RealSense camera.
    The therapist tablet was used by the experimenter, while participants only interacted with the user tablet.
    Each participant completed one session of the game.
    Before starting, each participant was provided with a verbal explanation of the game.
    Within a session, participants had to memorise randomly generated sequences of each difficulty level, such that there were game stages when no feedback was provided and stages when feedback was provided after solving consecutive sequences of the same length.
    The purpose was investigating the participants' performance in the game and engagement during the interaction with the robot, but also their reactions to the feedback given by the robot.
    For this purpose, for each participant, we collected the game performance, estimated engagement scores, and timings of recreating the sequences on the tablet.
    The collected data were used for learning user models and a behaviour model as described in section \ref{sec:behaviour_personalisation}.

    \subsection{Results}
    \label{sec:experiments:results}

\begin{figure}[tp]
        \centering
        \includegraphics[width=\linewidth]{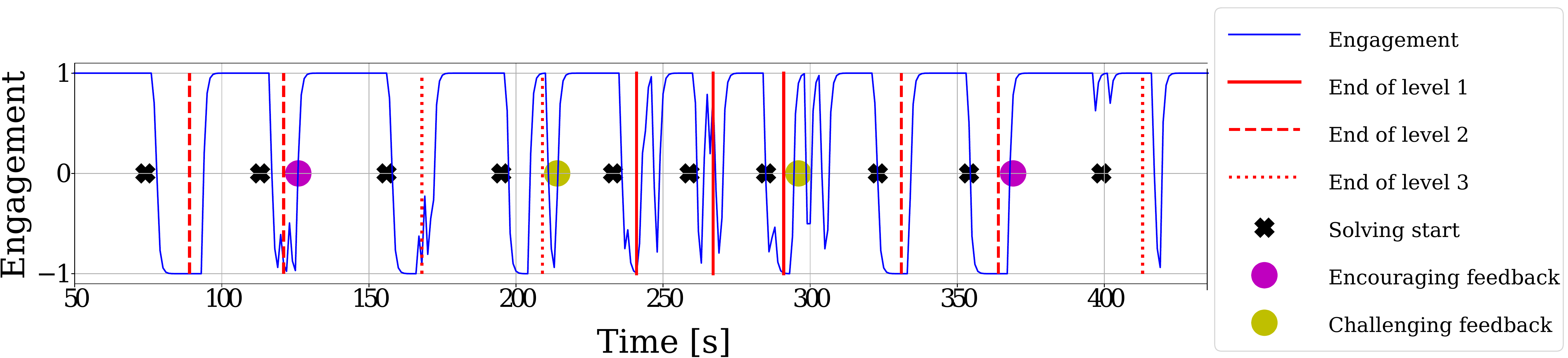}
        \caption{Expected $E$ in one session for one of the participants. The black crosses are times when the user had to recreate a sequence on the tablet; the red lines mark the ends of game stages (with a certain $L$), which are followed by either the start of a new sequence or robot feedback given to the user.}
        \label{fig:sequence_learning}
    \end{figure}

    An example evolution of the estimated engagement score during an interaction with one of the participants is presented in Fig. \ref{fig:sequence_learning}.
    It is important to mention that our system returns an engagement score ($1$ or $-1$) several times per second, but we make an assumption that a person's affective state would not significantly change within one second; thus, for all computations described in the previous sections, we use an \emph{expected engagement} value that is calculated for every second of the interaction.
    Fig.~\ref{fig:sequence_learning} presents data from a user whose engagement value is as expected, namely the engagement is generally high when the robot is talking to the user (after the participant finishes solving a sequence) and decreases when the participant is asked to recreate the sequence on the tablet (as they had to look down at the tablet).
    It is important to mention that some users did not behave as expected due to environmental disturbances or the way they were focusing on the robot; for instance, some users preferred to listen to the robot with closed eyes rather than look at it, which affects the engagement estimate.

\begin{figure}[tp]
		\begin{subfigure}[b]{0.19\textwidth}
			\includegraphics[width=\textwidth]{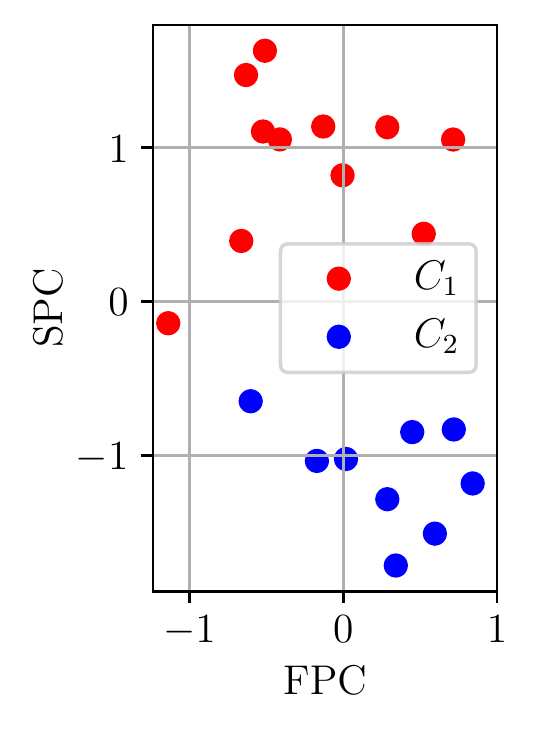}
			\subcaption{}%
			\label{subfig:clusters}
		\end{subfigure}
		\begin{subfigure}[b]{0.3\textwidth}
			\includegraphics[width=\textwidth]{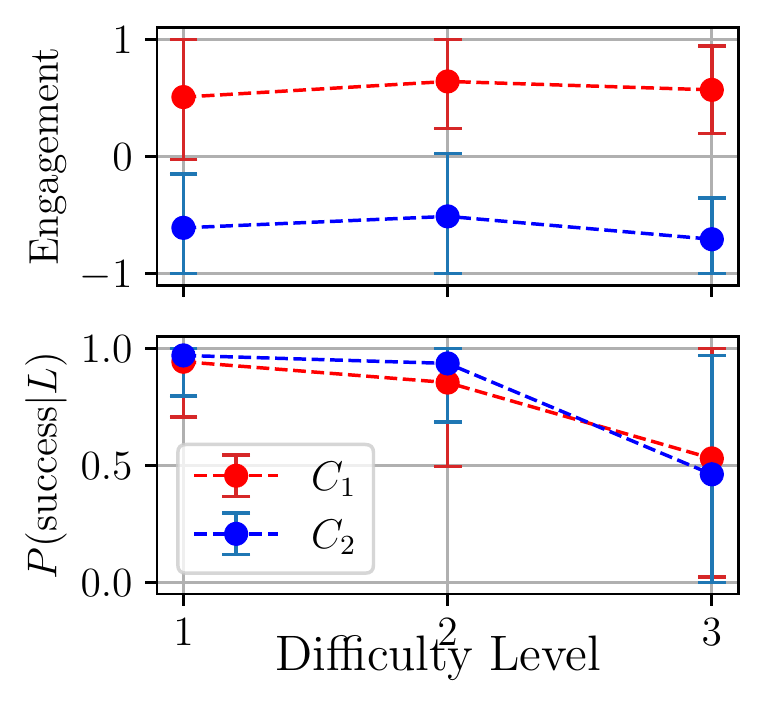}
			\subcaption{}%
			\label{subfig:clusters_statistics}
		\end{subfigure}
		\caption{(a) Obtained user clusters (FPC and SPC stand for first and second principal components, respectively). (b) Engagement level and probability of success given a certain difficulty level $L$ for each user cluster.}
		\label{fig:clustering_results}
	\end{figure}

    From the collected data, we created 20 vectors $\vec{u}_j$ as in Eq. \eqref{eq:user_vector}, projected them onto a 2D space, and clustered them into $c=2$ groups (Fig.~\ref{subfig:clusters}) using K-means clustering.
    We selected the number of clusters $c$ so that every cluster contains enough users to be further modelled as a GP; in our case, $|C_1| = 11$ and $|C_2| = 9$ participants, such that each cluster represents users that share a similar behaviour.
    For each difficulty level $L$ and user cluster $C_k$, Fig.~\ref{subfig:clusters_statistics} shows the mean and standard deviation values for the engagement $E$ and the probability of success given a certain difficulty level ($P(\text{success}|L)$).
    Based on the results, it can be seen that $C_1$ and $C_2$ are similar with respect to $P(\text{success}|L)$, but they significantly differ when it comes to $E$, as users belonging to $C_1$ show a much higher level of engagement in the interaction with the robot than those in $C_2$.
    After grouping the users, we trained four GP models, namely $F^p$ and $F^e$ for each cluster, and thus created two user models, $\mathcal{M}_1$ (for $C_1$) and $\mathcal{M}_2$ (for $C_2$).

\begin{figure}[tp]
		\centering
		\includegraphics[width=\linewidth]{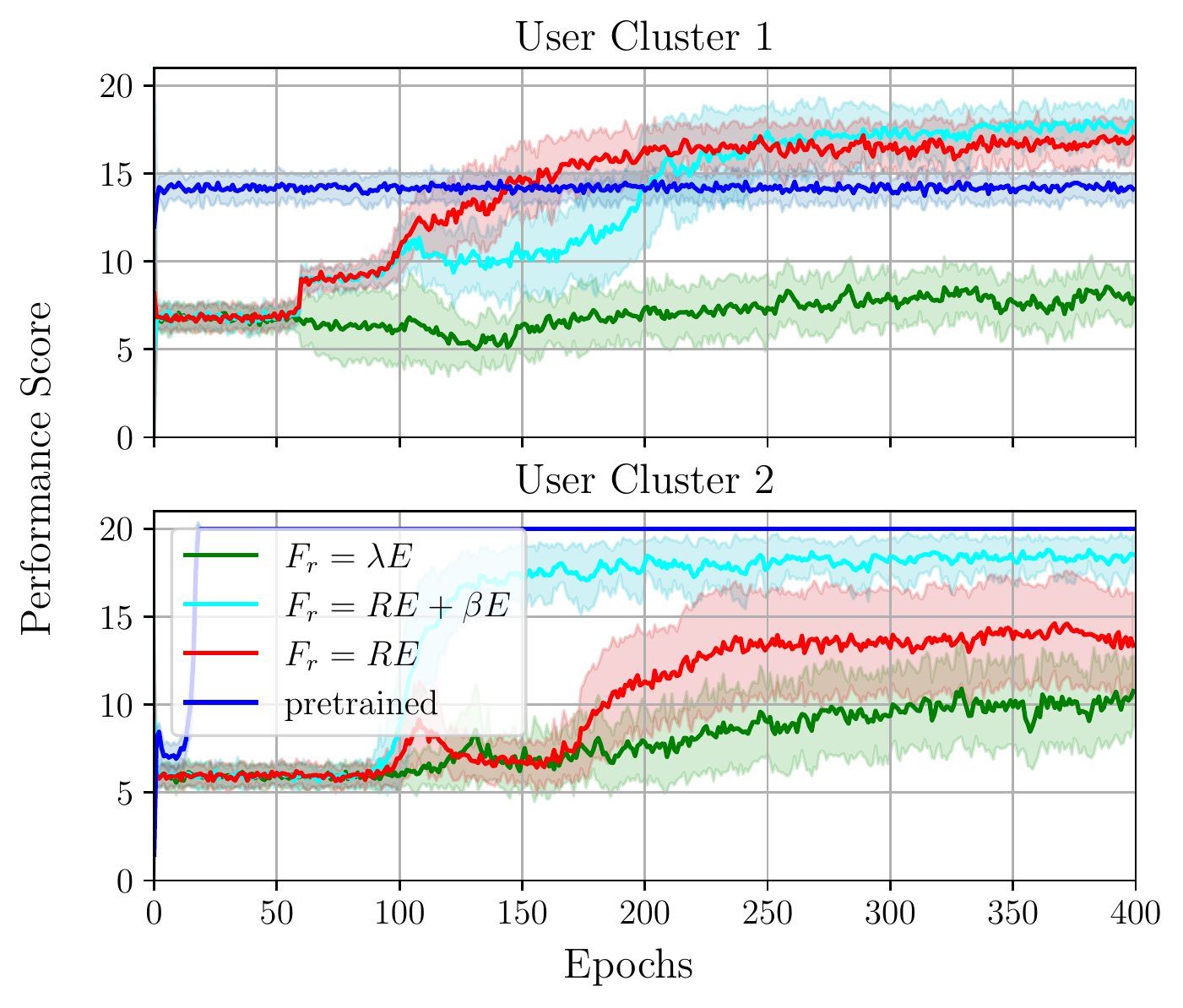}
		\caption{Comparison of different approaches for training behaviour models based on the user model performance score}
		\label{fig:update_pretrain_comparison_score}
	\end{figure}

\begin{figure}[tp]
		\centering
		\includegraphics[width=\linewidth]{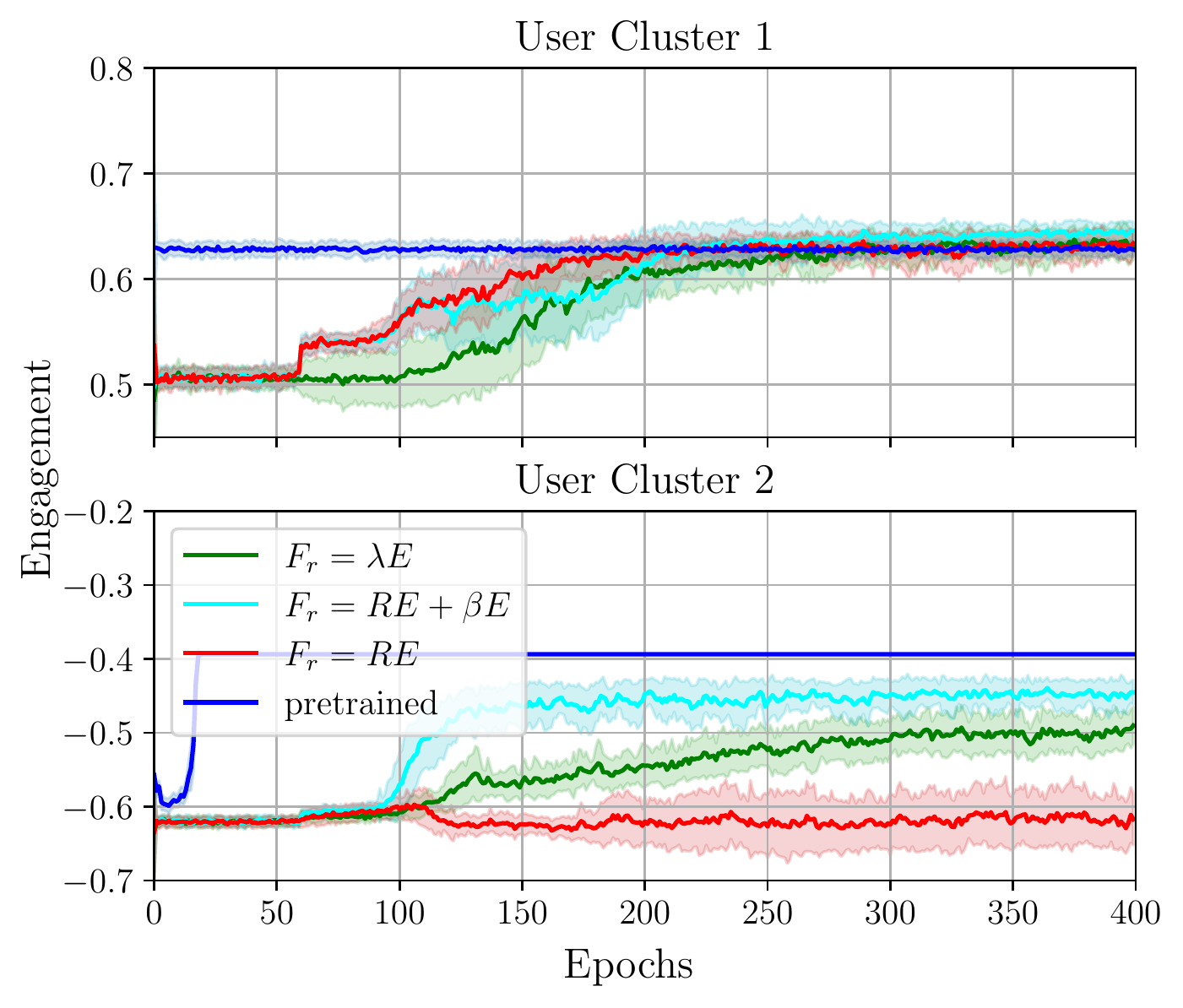}
		\caption{Comparison of different approaches for training behaviour models based on the user model engagement}
		\label{fig:update_pretrain_comparison_engagement}
	\end{figure}

	Given $\mathcal{M}_1$ and $\mathcal{M}_2$, we trained two behaviour models as described in section \ref{subsec:robot_behaviour_model}.
    To find the best method for calculating the reward $R_a(\vec{s}_t)$, we compared three different candidates for the function $F_r$, namely (i) $F_r = RE$, (ii) $F_r = RE + \beta E$, and (iii) $F_r = \lambda E$.
    We particularly aimed to check what influence $RE$ and $E$ have on the quality and speed of the policy convergence for both user models.
    The hyperparameters $\beta$ and $\lambda$ depend on the range of values of $RE$ and $E$ and were selected empirically; in our case, $\beta = \lambda = 3$.
    The average results of the training procedure (over 30 runs) are shown in Fig. \ref{fig:update_pretrain_comparison_score} and Fig. \ref{fig:update_pretrain_comparison_engagement}.
    In the figures, one training epoch is equal to $100$ sessions of a sequence learning game, where each session means that the user has to solve $\omega = 10$ sequences.
    The performance score (Fig. \ref{fig:update_pretrain_comparison_score}) stands for the mean accumulated score (accumulated in one session and averaged over the epoch).
    As shown in Fig. \ref{fig:update_pretrain_comparison_score}, calculating the reward by combining both $E$ and $RE$ helps in quick personalisation of the game difficulty for $\mathcal{M}_2$, but is not more advantageous in comparison to using only $RE$ for $\mathcal{M}_1$.
    On the other hand, when using $F_r = \lambda E$, the trained policy gives the worst results with respect to the performance score.
    We obtained different results when evaluating the training process with respect to the user's engagement.
    In Fig. \ref{fig:update_pretrain_comparison_engagement}, it can be noted that, for $\mathcal{M}_1$, all three versions of $F_r$ lead to similar results, while the lowest engagement for $M_2$ is obtained when the engagement information is ignored in the reward.
    The policy training seems to have better results when $F_r = \lambda E$ and can be meaningfully improved when both $RE$ and $E$ are considered.
    Based on the aforementioned results, it can be concluded that adjusting the task difficulty --- in our case, using the game score for computing $R_a(\vec{s}_t)$ --- can help in increasing the engagement; this is in line with \cite{tsiakas2016adaptive}.

	To increase the policy convergence speed, we also attempted policy transfer, namely the Q-table for one cluster was initialized with the Q-table that was trained on the other user model.
    The initial policy was chosen (out of $30$ learned policies) based on the highest average return value in the last pretraining epoch.
    Here, training was performed with $F_r = RE + \beta E$ and an exploitation-only-based strategy, as exploration might lead to undesired robot actions during real-life therapeutic scenarios.
    The results of this analysis are shown in Fig. \ref{fig:update_pretrain_comparison_score} and Fig. \ref{fig:update_pretrain_comparison_engagement}.
    Analysing both the performance score and engagement, it can be noticed that the initialisation of the lower engagement policy ($\mathcal{M}_2$'s) with the higher engagement policy ($\mathcal{M}_1$'s) improves not only the speed, but also the quality of personalisation, such that the initial policy can not only adapt, but it also gives better results than the user-specific policy.
    When initialising $\mathcal{M}_1$'s policy with $\mathcal{M}_2$'s policy, however, the engagement and performance score remain unchanged over the entire training procedure. As in \cite{tsiakas2016adaptive}, this may be caused by obtaining mainly positive rewards during training, as $\mathcal{M}_1$ usually outputs a positive $E$.

%% file: subfiles/discussion.tex
    \section{DISCUSSION AND CONCLUSIONS}
    \label{sec:discussion}
    In this work, we presented an approach for increasing the autonomy of a robot that can be used during therapy for children with ASD.
    We particularly described a robot behaviour model that can be used to learn personalised robot policies (in terms of provided feedback and activity difficulty level) for groups of similar users.
    We also created user models from data collected during a feasibility study with adult participants without ASD; these models were used for training personalised behaviour models.
    Based on the results from this experiment, we can conclude that computing rewards based on both user engagement and game performance score generally leads to faster policy convergence.
    We additionally performed policy transfer experiments; policy transfer can significantly improve the policy convergence speed, but may also lead to undesired results if the initialising policy or reward function are inappropriate.

    In future work, we need to improve the method for updating the Q-table so that the pretrained policy converges to the optimal one.
    The initial policy (even trained on the same user) also needs to be adapted efficiently and dynamically, as a user may change their characteristics over time, which may not be modelled with the current user model.
    It is also important to note that, although policy transfer can meaningfully improve the behaviour model, training the model is still a slow process, as the optimal policy seems to be reached only after $10$--$20$ epochs.
    We believe that this effect can be alleviated by applying the learning from guidance concept \cite{esteban2017build,tsiakas2016adaptive,sequeira2016,andriella2022}, which would also make the system safer and more applicable to real therapy.